\documentclass{article}

\usepackage{microtype}
\usepackage{booktabs} 

\usepackage[ruled,vlined]{algorithm2e}
\usepackage{authblk}
\usepackage{mdframed}
\usepackage{amsmath}
\usepackage{stmaryrd}
\usepackage{multirow}
\usepackage{wrapfig}
\usepackage[pdftex]{graphicx}
\usepackage{subfig}
\usepackage{xcolor}
\usepackage{colortbl}

\usepackage{hyperref}

\definecolor{Gray}{gray}{0.85}
\newcolumntype{a}{>{\columncolor{Gray}}l}
\newcommand{\std}[1]{\scriptsize{$\pm$#1}}

\newcommand{\xhdr}[1]{\vspace{0em}\noindent{{\bf #1.}}}
\newcommand{\name}{\textsc{aware}\xspace}

\definecolor{darkpastelgreen_cell}{HTML}{00a651}
\definecolor{darkpastelblue}{HTML}{00aeef}
\definecolor{darkpastelpink}{HTML}{e75480}

\usepackage[accepted]{icml2021}

\icmltitlerunning{Deep Contextual Learners for Protein Networks}

\begin{document}

\twocolumn[

\icmltitle{
Deep Contextual Learners for Protein Networks 
}




\begin{icmlauthorlist}
\icmlauthor{Michelle M. Li}{dbmi}
\icmlauthor{Marinka Zitnik}{dbmi,broad,hdsi}
\end{icmlauthorlist}


\icmlaffiliation{dbmi}{Harvard University}
\icmlaffiliation{broad}{Broad Institute of MIT and Harvard}
\icmlaffiliation{hdsi}{Harvard Data Science}

\icmlcorrespondingauthor{Michelle M. Li}{michelleli@g.harvard.edu}
\icmlcorrespondingauthor{Marinka Zitnik}{marinka@hms.harvard.edu}

\icmlkeywords{Machine Learning, ICML}

\vskip 0.3in
]



\printAffiliationsAndNotice{}  

\begin{abstract}
Spatial context is central to understanding health and disease. Yet reference protein interaction networks lack such contextualization, thereby limiting the study of where protein interactions likely occur in the human body and how they may be altered in disease. Contextualized protein interactions could better characterize genes with disease-specific interactions and elucidate diseases' manifestation in specific cell types.
Here, we introduce \name, a graph neural message passing approach to inject cellular and tissue context into protein embeddings. \name optimizes for a multi-scale embedding space, whose structure reflects network topology at a single-cell resolution. 
We construct a multi-scale network of the Human Cell Atlas and apply \name to learn protein, cell type, and tissue embeddings that uphold cell type and tissue hierarchies. 
We demonstrate \name's utility on the novel task of predicting whether a protein is altered in disease and where that association most likely manifests in the human body. To this end, \name outperforms generic embeddings without contextual information by at least $12.5\%$, showing \name's potential to reveal context-dependent roles of proteins in disease.
\end{abstract}

\section{Introduction}
Modeling interactions between proteins has been crucial for disentangling the rich diversity of complex biological phenomena, such as the mechanisms underlying organ function in multicellular organisms~\cite{luck2020reference,bassett2017network} and disease processes~\cite{cheng2018network, menche2015uncovering,sonawane2019network}. 
However, protein interaction (PPI) networks are typically presented as generic maps without contextual information about the particular cell type, disease state, and external stresses and stimuli on the system, which are known to have an influence on protein interactions and their function~\cite{przytycka2010toward,ideker2012differential,zhang2016construction,willsey2018psychiatric,markmiller2018context}.
As a result, there have been many efforts to contextualize protein interactions, which have only further revealed underlying mechanisms of disease progression and drug action \cite{basha2020differential, greene2015understanding, zitnik2017predicting}. Due to recent advancements in single-cell sequencing technology, we are now better equipped to investigate proteins and their roles in disease across tissues and at a single-cell resolution \cite{kamies2020advances, saviano2020single}.

Integrated single-cell transcriptomics and PPI network maps have already yielded insights into functional modules and gene-interaction dynamics of disease \cite{klimm2020functional, mohammadi2019reconstruction}. Prevailing approaches construct cell type specific PPI subnetworks and apply network analytical techniques on them individually \cite{cha2020single}. While such contextualized protein networks are powerful resources for descriptive analyses, they cannot generate optimized representations and provide predictions in downstream tasks. They could greatly benefit from the predictive capacity provided by deep graph learning, an emerging area founded upon key principles of network science \cite{li2021representation}. A deep graph learning approach could capture networks' topology and generate compact embeddings that could be easily specialized to cellular and tissue contexts. 

Here, we develop \name, a multi-scale graph neural network (GNN) to learn embeddings at various scales, guided by the hierarchical structure and function of cells and tissues. \name integrates cellular and tissue contextual information in the GNN architecture itself. As a result, \name can incorporate cell-specific gene expression into PPI networks while injecting the structure of protein interactions and cell type and tissue hierarchies. We evaluate \name on the Human Cell Atlas and cell type specific disease-gene associations, and demonstrate that \name embeddings outperform generic embeddings by at least 12.5\%, thus highlighting the importance of contextual learners for protein networks.

\section{Methods}
\begin{figure*}[ht]
    \centering
    \includegraphics[width=\textwidth]{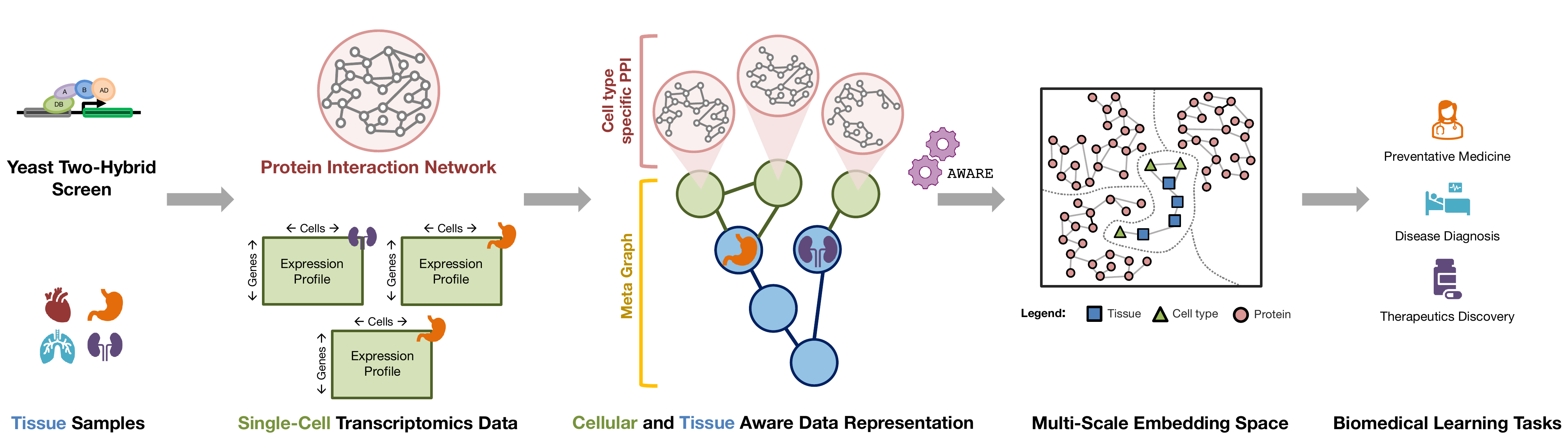}
    \caption{\textbf{Overview of the approach.} We construct cell type specific PPI subnetworks from a global PPI network (predominantly constructed by yeast two-hybrid screens) and single-cell transcriptomics data (usually generated from tissue samples). We inject cellular and tissue context based on single-cell transcriptomics meta-data and the tissue hierarchical ontology to create a data representation on which we apply \name to generate multi-scale protein, cell type, and tissue embeddings. Finally, we utilize such embeddings for downstream biomedical learning tasks related to preventative medicine, disease diagnosis, and therapeutics discovery.}
    \vspace{-3mm}
    \label{fig:overview}
\end{figure*}

We start by describing the construction of our cell type specific PPI subnetworks, and proceed with an overview of our multi-scale graph neural network approach, \name.

\subsection{Constructing cell type specific protein networks}
\label{method:celltype-ppi}

We extract cell type specific PPI subnetworks from a global PPI network based on differential gene expression without making any modifications. We first process, normalize, cluster, and annotate our single-cell transcriptomics (scRNA-seq) dataset using SCANPY \cite{wolf2018scanpy} and SCSA \cite{cao2020scsa}. We then select the top $k$ differentially expressed genes in each cluster (or cell type). We create a subgraph $S$ from these $k$ nodes, and continuously add $n$ nodes until $S$ is connected. Ultimately, each cell type specific PPI subnetwork $S_c$ (for $c \in C$ cell types) consists of the top $k + n$ differentially expressed genes.

\subsection{Contextualing embeddings of protein networks}
\label{method:algo}

Our method, \name, has four components: (1)~\textit{Initializing} cell type specific PPI embeddings using node- and semantic-level attention (Sec.~\ref{method:att}); (2)~\textit{Up-pooling} from cell type specific PPI embeddings to update cell type embeddings (Sec.~\ref{method:pool}); (3) \textit{Updating} cell type and tissue embeddings using node- and semantic-level attention (Sec.~\ref{method:att}); and (4) \textit{Updating} cell type specific PPI embeddings using node- and semantic-level attention, and \textit{down-pooling} from cell type embeddings to regularize the cell type specific PPI embeddings (Sec.~\ref{method:pool}). In the following sections, we explain the details of our node- and semantic-level attention, up- and down-pooling steps, and overall objective function.

\subsubsection{Node- and semantic-level attention}
\label{method:att}

We aggregate information from neighbors in each meta-path $\phi_r \in \Phi$ (i.e.~sequence of node types; see \citet{sun2011pathsim} for the definition of meta-paths) using node- and semantic-level attention mechanisms \cite{wang2019heterogeneous}.
Our \textit{node-level attention} learns the importance $\alpha^{\phi_r}_{i,j}$ of node $n_j$ to its neighbor $n_i$ connected by meta-path $\phi_r$ at layer~$l$: $\mathbf{z}^l_{n_i}(\phi_r) = \bigparallel_{k=1}^K \sigma \big( \sum_{j \in \mathcal{N}^{\phi_{r}}_i} \alpha^{\phi_{r}}_{i, j} \mathbf{M}^{\phi_r}\mathbf{h}^{l-1}_{n_j} \big)$, where $K$ is the number of attention heads, $\bigparallel$ denotes concatenation, $\sigma$ is the nonlinear activation function, $\mathcal{N}^{\phi_r}_i$ is the set of neighbors for $n_i$ connected by $\phi_r$ (includes itself via self-attention), $\mathbf{M}^{\phi_r}$ is a type-specific transformation matrix to project the features of $n_i$, and $\mathbf{h}^{l-1}_{n_j}$ is the previous layer's embedding for $n_j$. 
After generating node embeddings for every meta-path, our \textit{semantic-level attention} learns the importance $\beta^{\phi_r}$ of each meta-path $\phi_r \in \Phi$ to $n_i$ such that $\mathbf{h}^l_{n_i} = \sum_{\phi_r \in \Phi} \beta^{\phi_r} \mathbf{z}^l_{n_i}(\phi_r)$. We learn protein embeddings via the \textit{protein-protein} meta-path, and cell type and tissue embeddings via the \textit{celltype-celltype}, \textit{celltype-tissue}, and \textit{tissue-tissue} meta-paths.

\subsubsection{Up-pooling and down-pooling}
\label{method:pool}

Since we define cell types by their most differentially-expressed genes (Sec.~\ref{method:celltype-ppi}), we up-pool information from the embeddings of such genes in their corresponding PPI subnetworks. We initialize cell type embeddings by taking the average of their proteins' embeddings, weighted by the proteins' relative differential expression $d^c_i$ such that 
$\mathbf{h}^0_{c} = \sum_{i \in V_c} d^c_i \mathbf{h}^{0}_{n^c_i}$, where $\mathbf{h}^0_{n^c_i}$ is the embedding of node $n_i \in V_c$ in the PPI subnetwork for cell type $c$. We initialize tissue embeddings by taking the average of their neighbors: $\mathbf{h}^l_{t} = \frac{1}{|N_t|}\sum_{i \in N_t} \mathbf{h}^{l-1}_i$. For $l > 0$, we up-pool from cell type specific PPI embeddings using an attention mechanism to learn the importance $\gamma^c_i$ of node $n_i$ to cell type $c$ such that $\mathbf{h}^l_{c} = \mathbf{h}^{l-1}_{c} + \bigparallel_{k=1}^K \sigma \big( \sum_{i \in V_c} \gamma^c_i \mathbf{h}^{l-1}_{n^c_i} \big)$. Note that we are learning the weights of proteins that are important, which could be useful for identifying novel cell type biomarkers.

After propagating cell type and tissue information in the meta graph via node- and semantic-level attention (Sec.~\ref{method:att}), we down-pool from cell type $c$ to its corresponding PPI subnetwork by leveraging the learned importance $\gamma^c_i$ of node $n_i$ to $c$ such that $\mathbf{h}^l_{n^c_i} = \mathbf{h}^l_{n^c_i} + \gamma^c_i\mathbf{h}^l_{c}$. Intuitively, we are imposing the structure of the meta graph onto the PPI subnetworks based on the proteins' importance to their corresponding cell type's identity.

\subsubsection{Overall objective function}
\label{method:loss}

To capture the structure of cell type specific PPI subnetworks and the meta graph, we apply softmax $\mathcal{L}_S$ on link prediction. Additionally, to optimize meta-graph-based clustering of the cell type specific PPI embeddings, we use center loss \cite{wen2016discriminative} to predict the associated cell type of each protein: $\mathcal{L}_C = \frac{1}{2}\sum^m_{i=1} || \mathbf{x}_i - \mathbf{c}_{y_i} ||^2_2$ for $m$ samples. Thus, $\mathcal{L} = \mathcal{L}_S + \lambda\mathcal{L}_C$, where $\lambda$ is a hyperparameter.

\section{Experimental setup and results}
Here, we describe our protein, cellular, and tissue datasets, baseline methods, and implementation details. 

\subsection{Datasets}\label{sec:data}

\xhdr{Global protein interaction network}
Our global PPI network is the union of physical multi-validated interactions from BioGRID \cite{oughtred2019biogrid, stark2006biogrid}, the Human Reference Interactome \cite{luck2020reference}, and \cite{menche2015uncovering} with 15,461 nodes and 207,641 edges.

\xhdr{Human Cell Atlas}
We integrate seven 10X scRNA-seq datasets from healthy individuals in the Human Cell Atlas (HCA) \cite{regev2017science}, and obtain 64 cell type specific PPI subnetworks, each with 1,645 proteins on average.

\xhdr{Cell type and tissue hierarchies}
The meta graph is composed of cell type and tissue nodes. Cell type nodes are defined by the annotated clusters from scRNA-seq analysis (Sec.~\ref{method:celltype-ppi}); cell-cell interactions, by CellPhoneDB \cite{efremova2020cellphonedb}; celltype-tissue relationships, by meta-data from HCA; and tissue hierarchy, by BRENDA Tissue Ontology \cite{gremse2010brenda}. Here, we use 8 (of 64) cell types and 4 (of 50) tissues from the meta graph (Fig.~\ref{fig:res-embed}a).

\xhdr{Disease-gene associations at single-cell resolution}
We extract single-cell disease-gene associations from SC2disease \cite{zhao2021sc2disease} for breast, liver, and lung cancers, Alzheimer's disease (AD), rheumatoid arthritis (RA), autism spectrum disorder (ASD), lupus nephritis, pancreatic ductal adenocarcinoma, multiple sclerosis (MS), and atherosclerosis. Among these diseases, the overlapping cell types are T cells, astrocytes, and monocytes. The global and cell type specific PPI and our SC2disease dataset share 245 genes. We train and evaluate a KNN for predicting cell type specific disease-gene associations on 80\% and 20\% of the genes, respectively (Tbls.~\ref{tab:knn}-\ref{tab:per-label}). In this limited dataset, only 20 labels have enough samples for evaluation.

\subsection{Baseline methods and implementation details}

To demonstrate the necessity of each component in \name, we perform ablation studies on its core features: cell type specificity (Sec.~\ref{method:celltype-ppi}), meta graph regularization (Sec.~\ref{method:pool}), and prototypical loss (Sec.~\ref{method:loss}). In \textsc{global}, we ignore scRNA-seq information; in \textsc{-mg}, we turn off down-pooling (i.e.~remove notion of cellular and tissue hierarchies); in \textsc{-proto}, we set $\lambda = 0$ (i.e.~do not optimally separate cell type specific protein embeddings). Across all methods, our node feature matrix has dimension 2048; output layer has dimension 128; node- and semantic-level attentions have 8 heads; softmax $\text{lr} = 0.001$ and center loss $\text{lr} = 0.01$. In \name, $\lambda=0.001$. We use PyTorch Geometric, Adam optimizer, and grid-search for hyperparameter tuning.

\subsection{Results}
Finally, we discuss the results of our experiments to evaluate the quality and utility of \name's embeddings.

\begin{figure*}
    \centering
    \includegraphics[width=\textwidth]{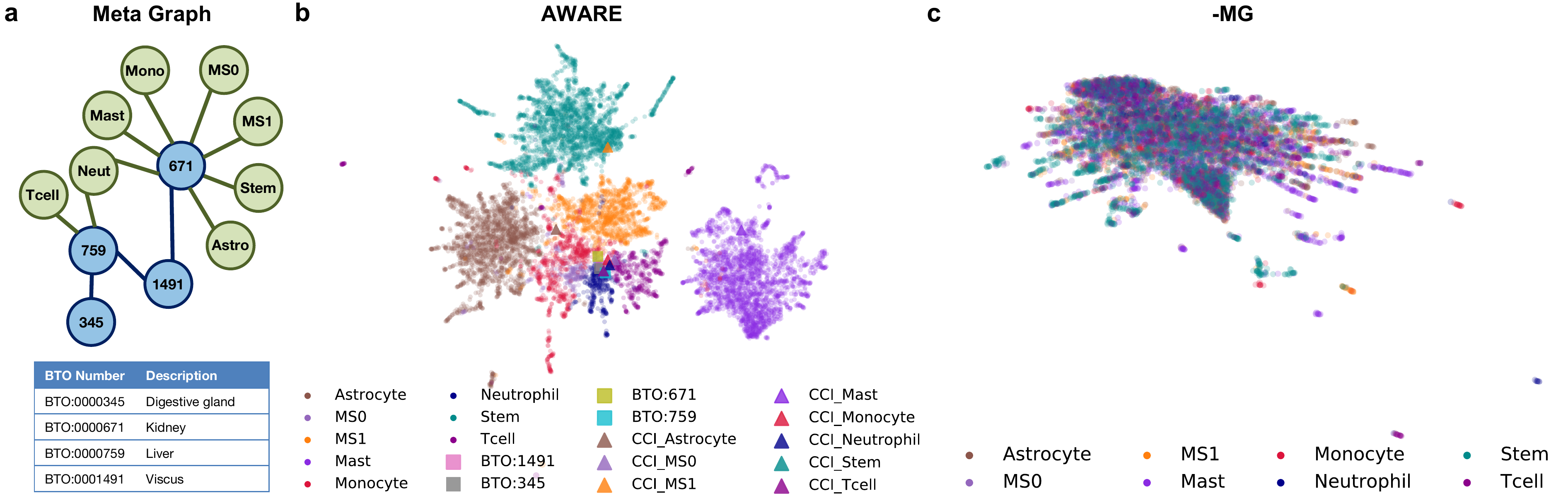}
    \caption{\textbf{Visualization of contextual embeddings by \name and alternative approaches.} Shown are 2D UMAP projections of our protein (circle), cell type (triangle), and tissue (square) embeddings from the \textbf{(b)} \name and \textbf{(c)} \textsc{-mg} methods. Unlike \textsc{-mp} (and \textsc{global}, not shown due to space constraints), \name (and \textsc{-proto}, also not shown) clusters according to the cellular and tissue hierarchies imposed by the \textbf{(a)} meta graph. Notably, cell type embeddings (triangle) are proximal to their corresponding clusters of protein embeddings (circle). BTO, BRENDA Tissue Ontology; CCI, cell-cell interaction; MS0, Mesenchymal Stem Cell 0; MS1, Mesenchymal Stem Cell 1.}
    \label{fig:res-embed}
\end{figure*}

\xhdr{Learning cellular and tissue aware embeddings}
Unlike \textsc{global} and \textsc{-mg} (Fig.~\ref{fig:res-embed}b), which are unable to discern cell type specificity in the embedding space, the cellular and tissue aware embeddings generated by \textsc{-proto} and \name (Fig.~\ref{fig:res-embed}c) cluster according to the meta graph's structure. While \textsc{-proto} has less clear cell-type-separation, it is notable that \textsc{-proto} and \name can generate highly granular embeddings that capture cell type and tissue hierarchies.

\xhdr{Predicting cell type specific disease-gene associations}
Embeddings generated with some notion of cellular and tissue hierarchy, i.e.~\name and \textsc{-proto}, outperform those without it by at least 12.5\% (Tbl.~\ref{tab:knn}). \name and \textsc{-proto} embeddings have comparable or better performance than \textsc{global} embeddings in 15 (of 20) labels (Tbl.~\ref{tab:per-label}). In particular, they outperform \textsc{global} in predicting genes in T cells associated with breast cancer and MS by up to 50\%; in monocytes associated with RA by 200\%; and in astrocytes associated with atherosclerosis by 50\%. Such improvements highlight the need to contextualize protein networks with biologically meaningful cell type and tissue dependencies. 

\begin{table*}
\vspace{-8mm}
\caption{\textbf{Performance of protein embeddings on downstream prediction of disease-gene associations in 3 cell types.} Reported are average Micro-F1 scores at $k$ nearest neighbors, together with standard errors across 10 independent runs. Remarkably, as the number of neighbors $k$ increases, \textsc{global} and \textsc{-mg}'s performance decrease, likely due to poor separation of cell type specific protein embeddings.}
\label{tab:knn}
\renewcommand{\arraystretch}{0.7}
\begin{center}
\begin{small}
\begin{sc}
\begin{tabular}{cccc}
\toprule
method & $k=1$ & $k=5$ & $k=10$ \\
\midrule
global & 0.24\std{$<$0.01} & 0.20 \std{$<$0.01} & 0.06 \std{$<$0.01} \\
\textsc{-mg} & 0.15 \std{$<$0.01} & 0.13 \std{$<$0.01} & 0.02 \std{$<$0.01} \\
\textsc{-proto} & \textbf{0.27 \std{$<$0.01}} & \textbf{0.28 \std{$<$0.01}} & 0.11 \std{$<$0.01} \\
\name & \textbf{0.27 \std{$<$0.01}}  & 0.23 \std{$<$0.01} & \textbf{0.12 \std{$<$0.01}} \\
\bottomrule
\end{tabular}
\end{sc}
\end{small}
\end{center}
\vskip -0.1in
\end{table*}

\begin{table*}
\vspace{-8mm}
\caption{\textbf{Evaluation of protein embeddings on downstream prediction task.} Reported are average Recall@20 values and standard errors across 10 independent runs. A protein can be associated with 0, 1, or more diseases, with each association existing in the context of a specific cell type. The task is to predict cell type specific disease-gene associations from the learned protein embeddings. For instance, according to row \#1, a Recall@20 value of 0.8 means that 80\% of the genes truly associated with breast cancer when expressed in T cells are among the top 20 predicted by \name. Note that \name has slightly lower performance than \textsc{-proto} and \textsc{-mg} in two labels, likely due to the maximization of inter-class variation in \name, which can potentially separate similar proteins too much for the KNN to detect.}
\label{tab:per-label}
\renewcommand{\arraystretch}{0.7}
\begin{center}
\begin{small}
\begin{sc}
\begin{tabular}{lll|llla}
\toprule
disease & cell type & n\textsubscript{total} & global & \textsc{-mg} & \textsc{-proto} & \name \\
\midrule
breast cancer & \textcolor{darkpastelgreen_cell}{t cell} & 24 & 0.60 \std{0.16} & 0.90\std{0.10} & 0.90\std{0.10} & 0.80\std{0.13} \\
  & \textcolor{darkpastelpink}{astrocyte} & 2 & 0.20\std{0.13} & 0.20\std{0.13} & 0.20\std{0.13} & 0.20\std{0.13} \\
\midrule
liver cancer & \textcolor{darkpastelgreen_cell}{t cell} & 82 & 0.80\std{0.13} & 0.50\std{0.17} & 0.50\std{0.17} & 0.50\std{0.17} \\
  & \textcolor{darkpastelpink}{astrocyte} & 7 & 0.20\std{0.13} & 0.30\std{0.15} & 0.30\std{0.15} & 0.30\std{0.15} \\
\midrule
lung cancer & \textcolor{darkpastelgreen_cell}{t cell} & 117 & 0.90\std{0.10} & 0.60\std{0.16} & 0.60\std{0.16} & 0.60\std{0.16} \\
  & \textcolor{darkpastelpink}{astrocyte} & 8 & 0.40\std{0.16} & 0.40\std{0.16} & 0.40\std{0.16} & 0.40\std{0.16} \\
\midrule
alzheimer's disease (ad) & \textcolor{darkpastelgreen_cell}{t cell} & 37 & 0.50\std{0.17} & 0.10\std{0.10} & 0.00\std{$<$0.01} & 0.20\std{0.13} \\
  & \textcolor{darkpastelpink}{astrocyte} & 317 & 1.00\std{$<$0.01} & 1.00\std{$<$0.01} & 1.00\std{$<$0.01} & 1.00\std{$<$0.01} \\
  & \textcolor{darkpastelblue}{monocyte} & 10 & 0.10\std{0.10} & 0.20\std{0.13} & 0.20\std{0.13} & 0.20\std{0.13} \\
\midrule
rheumatoid arthritis (ra) & \textcolor{darkpastelgreen_cell}{t cell} & 105 & 1.00\std{$<$0.01} & 0.80\std{0.13} & 0.90\std{0.10} & 0.90\std{0.10} \\
  & \textcolor{darkpastelpink}{astrocyte} & 15 & 0.10\std{0.10} & 0.30\std{0.15} & 0.30\std{0.15} & 0.30\std{0.15} \\
  & \textcolor{darkpastelblue}{monocyte} & 71 & 0.20\std{0.13} & 0.60\std{0.16} & 0.60\std{0.16} & 0.60\std{0.16} \\
\midrule
autism spectrum disorder (asd) & \textcolor{darkpastelgreen_cell}{t cell} & 12 & 0.40\std{0.16} & 0.40\std{0.16} & 0.10\std{0.10} & 0.10\std{0.10} \\
  & \textcolor{darkpastelpink}{astrocyte} & 206 & 1.00\std{$<$0.01} & 1.00\std{$<$0.01} & 1.00\std{$<$0.01} & 0.90\std{0.10} \\
\midrule
pancreatic ductal adenocarcinoma & \textcolor{darkpastelgreen_cell}{t cell} & 30 & 0.30\std{0.15} & 0.80\std{0.13} & 0.70\std{0.15} & 0.60\std{0.16} \\
\midrule
multiple sclerosis (ms) & \textcolor{darkpastelgreen_cell}{t cell} & 90 & 0.60\std{0.16} & 0.70\std{0.15} & 0.50\std{0.17} & 0.70\std{0.15} \\
  & \textcolor{darkpastelpink}{astrocyte} & 239 & 1.00\std{$<$0.01} & 1.00\std{$<$0.01} & 1.00\std{$<$0.01} & 1.00\std{$<$0.01} \\
  & \textcolor{darkpastelblue}{monocyte} & 4 & 0.10\std{0.10} & 0.20\std{0.13} & 0.20\std{0.13} & 0.20\std{0.13} \\
\midrule
atherosclerosis (ath.) & \textcolor{darkpastelgreen_cell}{t cell} & 544 & 1.00\std{$<$0.01} & 1.00\std{$<$0.01} & 1.00\std{$<$0.01} & 1.00\std{$<$0.01} \\
  & \textcolor{darkpastelpink}{astrocyte} & 23 & 0.40\std{0.16} & 0.60\std{0.16} & 0.60\std{0.16} & 0.60\std{0.16} \\
\bottomrule
\end{tabular}
\end{sc}
\end{small}
\end{center}
\vskip -0.1in
\end{table*}

\section{Discussion and conclusion}
To inject cellular and tissue context into biological network embeddings, we have developed \name, a methodology to generate rich embeddings of proteins, cell types, and tissues that adhere to cell type and tissue hierarchies. We show \name's utility on the novel task of predicting cell type specific disease-gene associations, demonstrating the importance of cellular and tissue context in protein networks. We envision \name to open up new possibilities for contextually adaptive embeddings in biomedicine.

\section*{Acknowledgements}

We thank our reviewers for their time and insights. M.M.L.~is supported by T32HG002295 from the National Human Genome Research Institute and a National Science Foundation Graduate Research Fellowship. We also gratefully acknowledge the support of NSF under nos.~IIS-2030459 and IIS-2033384, the Harvard Data Science Initiative, the Amazon Research Award, and the Bayer Early Excellence in Science Award.



\bibliography{references}
\bibliographystyle{icml2021}

\end{document}